# A Novel Method for the Recognition of isolated Handwritten Arabic Characters

Ahmed Sahlol[1] and Cheng Suen [2]

[1]Department of Computer Teacher preparation, Damietta University, Damietta, Egypt
asahlol@encs.concordia.ca
[2]Department of Computer Science, Concordia University, Canada
parmidir@encs.concordia.ca

## ABSTRACT

*There are many difficulties facing a handwritten Arabic recognition system such as unlimited variation in human handwriting, similarities of distinct character shapes, interconnections of neighbouring characters and their position in the word. The typical Optical Character Recognition (OCR) systems are based mainly on three stages, preprocessing, features extraction and recognition. This paper proposes new methods for handwritten Arabic character recognition which is based on novel preprocessing operations including different kinds of noise removal also different kind of features like structural, Statistical and Morphological features from the main body of the character and also from the secondary components. Evaluation of the accuracy of the selected features is made. The system was trained and tested by back propagation neural network with CENPRMI dataset. The proposed algorithm obtained promising results as it is able to recognize 88% of our test set accurately. In Comparable with other related works we find that our result is the highest among other published works.*

## 1. INTRODUCTION

The Arabic alphabet is used by a wide variety of languages besides Arabic (especially in Africa and Asia) such as Persian, Kurdish, Malay and Urdu. The estimated number of historical Arabic manuscripts exceeds three millions [1], so we translate images of typewritten or handwritten text into machine-editable text encoded in a standard encoding scheme (ASCII). Handwritten text recognition of such as Arabic text is an active research problem [2].

OCR systems have expanded to recognize Latin alphabets, Japanese Katakana syllabic characters, Kanji (Japanese version of Chinese) characters, Chinese characters, Hangul characters, etc.

Work on Arabic OCR started in 1970s [3]. The first published work on Arabic OCR dates back to 1975 [5]. The first Arabic OCR system was made available in 1990s [6]. The recognition of Arabic handwriting presents some unique challenges and benefits to the researchers [7]. Although more than three decades have passed, there has been a lack of effort in the recognition of Arabic handwritten texts compared to the recognition of texts in other scripts [6].

There are many other applications for analysis of human handwriting such as writer identification and verification, form processing, interpreting handwritten postal addresses on envelopes and reading currency amounts on bank checks etc. The main problem encountered when dealing with handwritten Arabic characters is that characters written by different persons representing the same character are not identical but can vary in both size and shape. Unlimited variation in human handwriting styles similarities of distinct character shapes, character overlaps, and interconnections of neighboring characters. In addition, the mood of the writer and the writing situation can have an effect on writing styles.

Arabic Writing System is different than English; Arabic is written from right to left and is always cursive. It has 28 basic characters. Thus, roughly the alphabet set can expand to 84 different shapes according to the position of the letter (beginning, middle, end or isolated) as well as according to the style of writing (Nasekh, Roqa'a, Farisi and others).

A character is drawn in an isolated form when it is written alone and is drawn in three other forms when it is written connected to other characters in the word. For example, the character Ain has four forms: isolated (ع), initial (عـ), medial (ـعـ) and final (ـع).

The secondary components are character components that are disconnected from the main body.

Sixteen Arabic letters have from one to three secondary components (dots).

The type and position of the secondary components are very important features of Arabic letters. For example, Tah (ط) and Thah (ظ) differ only by the number of dots above the main body, Seen (س) and Sheen (ش), Saad (ص) and Daad (ض).

Another important kind of variations in drawing the secondary components appears mostly in drawing two or three dots. As shown in table1 Sheen (ش), the three dots come in three variations: isolated, connected and linked to the character, respectively. Also Taa (ت) can be drawn in two variations: two isolated dots or one short horizontal dashed line.

Another kind of differences between characters depended only on the position of the two dots; Taa (ت), Yaa (ي) and Taa (ت), Yaa (ي) respectively.

There is also another classification challenge; that some characters which contain secondary components can also be written without those secondary components depending on "Roqa'a" writing styles. For example is Alif (أ) which can be drawn with hamza (default) or without it (ا).

Another difficulty in recognizing the secondary components comes due to quickly writing, as writers draw them connected to the main body.

Intensive researches have been done to solve the problem of handwritten Arabic character recognition. Various approaches have been proposed to deal with this problem. Many approaches have been adopted in various ways to improve accuracy and efficiency.

Gheith et al. [7] extracted 96 features from the letter's secondary components, main body, skeleton, and boundary. These features are evaluated and best subsets of varying sizes are selected using five feature selection techniques. The evolutionary algorithm has the highest time complexity but it selects feature subsets that give the highest recognition accuracies.

Abdelazeem et al [8] and Abandah et al [9] used vertical and horizontal projections which gave us more valuable information to capture the distribution of ink along one of the two dimensions in the character. Another kind of useful feature is topological features.

Many approaches and techniques have been proposed like [10-14] used loops, dots, curves, relative locations, height, sizes of parts of characters, loop positions and types, line positions and directions and turning points. Others like [15-17] used statistics from moments of horizontal and vertical projection. Histogram of slopes along contour is used by [18].

Recent attempts for online recognition of Arabic characters can be seen in Kherallah et al. [19],[20], Mezghani and Mitiche [21], Saabni and El-Sana [22], and Sternby et al.[23].

Benjelil et al. [24], Ben Cheikh et al. [25], Kanoun et al. [26], Khan et al. [27], Ben Moussa et al. [28], Prasad et al. [29], Saeeda and Albakoor [30], and Slimane et al. [31] work with printed Arabic text.

In this paper we use Neural Networks as a classifier like [32] which proposed a system for the recognition of handwritten Arabic characters recognition. Also Sherif and Mostafa [33-43] presented a parallel design for backpropagation Neural Networks approach in order to accelerate the computation process. But another kind of Neural network called Learning Vector Quantization (LVQ) was used in [35] for handwritten Arabic character recognition.

Bluche and Ney [36], Kozielski [37] made a combination of a convolutional neural network with a HMM gave better results compared with recurrent neural networks, instead of using only HMM in [38]. Another combination between neural networks and HMM had been done in [39]. While others [40-41] used SVM (support vector machine) as a classifier for Arabic numbers and for texts [42].

The goal of this work is to develop a reliable offline OCR system for handwritten Arabic characters.

First we make different kind of noise removal then we used different kind of features:

- Whole body features: Which represent features like vertical and horizontal projection and number of connected components like Hamza and dots.

- Secondary component features. Which represent features like secondary component height, width and the pixel ratio.

Back propagation Neural network is then used to classify the characters based on the features that were extracted from the input character. Figure 1 summarizes the methodology adopted in this paper.

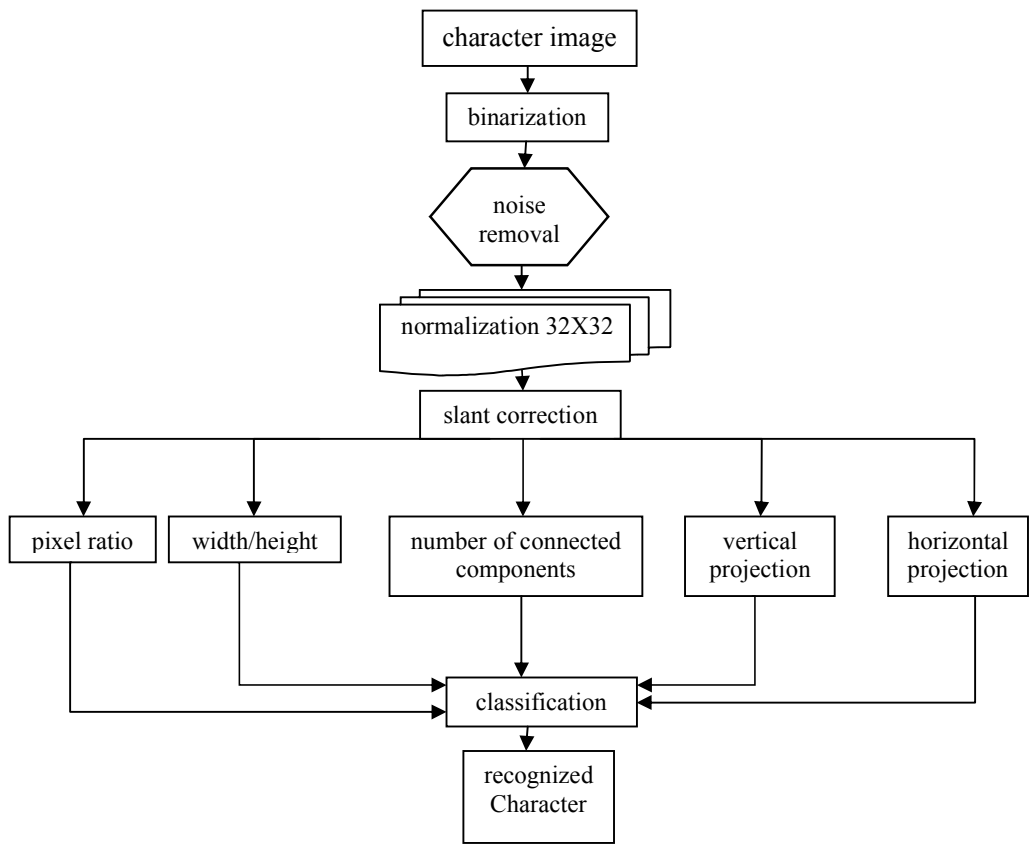

Figure1. Block diagram of the proposed system

This paper is organized in four sections. Section 2 describes the methodology of our work including Binarization, normalization, noise removal Algorithms, feature extraction techniques, Classification stage; which used classifier including its architecture, training and testing phase.

Section 3 provides experimental results included classification accuracy, comparison between our work and others and our contributions. Finally, Section 4 describes the main conclusions and future work.

## 2. MATERIALS AND METHODS

### 2.1 Binarization

Our purpose of this step is to convert the input image to binary image based on threshold. We use Otsu's method [43] to convert the grey character image to a binary image which is a normalized intensity value that lies in the range [0, 1].

we use Otsu method because it is a global binarization technique and also its short running time; the Mean running time for the Otsu's binarization method was 2.0 secs and this is one of the lowest running time (Original Sauvola algorithm [44] takes 12.6 secs).

### 2.2 Slant Correction

The basic idea is to locate near-vertical strokes in the character and estimate the average slant of the character from these strokes. Then, the slant in a character is corrected by applying a shear transformation to the character.

### 2.3 Normalization.

Normalization is to regulate the size, position, and shape of character images, so as to reduce the shape variation between the images of same class. Handwritten character images typically shows strong variability in appearance due to different writing styles.

Such variability calls for preprocessing techniques suitable for recognition of handwritten text. Size normalization is an important pre-processing technique in character recognition because the character image is mapped onto a predefined size so as to give a representation of fixed dimensionality for classification.

We use the Linear Backward mapping method [45] for binarization.

### 2.4 Noise removal

#### 2.4.1 Median filtering

Although that noise removal techniques have the effect of slightly distorting the actual image, but often this is a small price to pay for the removal of distracting noise and also we were so circumspect when choosing suitable techniques and their parameters. We use median filtering [46] for reducing random noise. For example, the image in figure2 (a) its middle pixel is affected by image noise, and has a value (100) much higher than the brightness values of the grey. If we take the pixel values and sort them in increasing order, we get: 25, 30, 30, 35, 35, 40, 40, 45, and 100. The middle of these nine values is 35, so we assign the value 35 to the middle pixel. The result looks like the second image (b). But by taking the middle (or median) value, the influence of the bright pixel is ignored. When we apply this sort of median filter all over an image with scattered pixels of noise, we effectively get rid of the noise.

| 25 | 30 | 35 | | 25 | 30 | 35 |
|----|-----|----|---|----|----|----|
| 30 | 100 | 40 | | 30 | 35 | 40 |
| 35 | 40 | 45 | | 35 | 40 | 45 |

Figure 2. Median filtering **(a)** Original image **(b)** Image after median filtering

In this paper, we adopt a 3 × 3 median filter was because it gave us the best result.

### 2.4.2 Morphological noise removal

**Filling**: fill isolated interior pixels, such as the center pixel in the pattern below see Figure 4 (a). For each pixel p in the binary image I, we check each pixel neighbors and decide whether P to be 0 or 1 if B(p) =7, where B(p) is the number of non-zero neighbors of p.

**Cleaning**: Remove isolated pixels, such as the centre pixel in the pattern below see Figure 4 (b). For each pixel p in the binary image I, we check each pixel neighbors and decide whether P to be 0 or 1, if B(p) =0, where B(p) is the number of zero neighbors of p.

**Removing**: Remove isolated pixels (individual 1s that are surrounded by 0s), such as the center pixel in the pattern below see Figure 4 (c). For each pixel p in the binary image I, check the four neighbors and decide whether P to be 0 or 1, if B(p) =0, where B(p) is the number of non-zero neighbors of p.

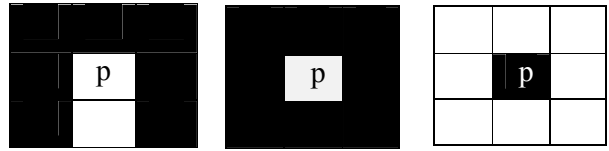

Figure 3. **(a)** filling     **(b)** cleaning     **(c)** removing

Another kind Morphological operation used in this paper is Dilation:

### 2.4.3 Dilation[47]

This Morphological technique exposes an image to a small shape called structuring element; this structuring element is positioned at all possible locations in the image and it is compared with the corresponding neighborhood of pixels. Some operations test whether the element "fits" within the neighborhood, while others test whether it "hits" or intersects the neighborhood. Figure 3 shows how dilation works.

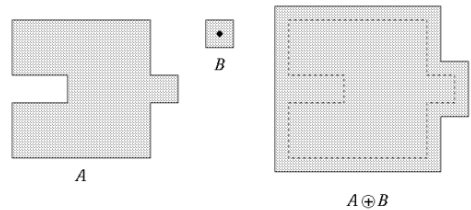

Figure 3. Exposing image A to a structuring element B

In this paper we use a square of 2x2 of ones as a structuring element as it gives us the best dilation job.

## 2.5 Feature Extraction

### 2.5.1 Structural Features:

### 2.5.1.1 Upper and Lower profile.

The upper and lower profiles capture the outlining shape of a connected part of the character. Upper (or lower) connected part profile is computed by measuring the distance (pixel count) of each column group from the top (or, bottom) of the bounding box of the connected part to the closest ink pixel in that column group.

### 2.5.1.2 Horizontal and Vertical projection profiles.

Vertical profile is the sum of white pixels perpendicular to the y axis. It is computed by scanning the character column wise along the y-axis and counting the number of white pixels in each column. The character is traced vertically along the y-axis. The column wise sum of number of white pixels present in each column.

Similarly, the horizontal projection profile is sum of black pixels but it is perpendicular to the x axis. The character is traced horizontally along the x-axis. The row wise sum of number of white pixels present in each row.

### 2.5.2 Statistical Features:

#### 2.5.2.1 Connected components.

A pixel p at coordinates (x,y) has adjacent neighbors whose coordinates are (x+1,y), (x-1,y), (x,y+1) and (x,y-1). This set of 4 neighbors of p denoted N4 (p) is illustrated in figure 4, where the four diagonal neighbors of p have coordinates (x+1,y+1), (x+1,y-1), (x-1,y+1) and (x-1,y-1) respectively.

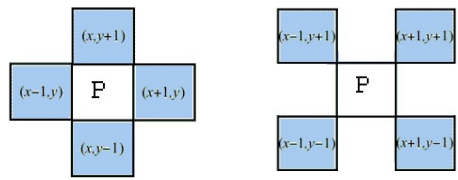

Figure 4.  **(a)** Adjacent neighbors  **(b)** Diagonal neighbors

Connected component is an important feature because most of Arabic characters contain one or more connected components like (ي ,ك ,ق ,غ ,ض ,ذ ,خ ,ث ,ت ,ب ,أ), so this feature can be effective for distinguishing among those characters which contain connected components and others which don't contain connected components.

We identify the main body easily as it is usually the largest component so any other connected components are considered as secondaries.

### 2.5.3 Topological features:

#### 2.5.3.1 End points.

Another feature that is useful is the number of end points in the character. Those points have only one neighbor and the other three neighbors are noise.

We dilate the image to join all its parts. Dilation essentially just adds pixels around existing white pixels, then we shrink the image to points (removes pixels so that objects without holes shrink to a point, and objects with holes shrink to a connected ring halfway between each hole and the outer boundary). Finally we find the pixel that is still white, which i is the row and j is the column of the white pixel in image such that image(i,j) is equal to 1. All the other pixels should be 0 in the image.

#### 2.5.3.2 Pixel Ratio.

The character area is the total number of white (foreground) and black pixels (background) of the character. Pixel ratio is: (the number of white pixels / the number of black pixels).

#### 2.5.3.3 Height to Width Ratio.

Since different people write same characters in different sizes, the absolute width and height are not reliable features for Arabic handwritten characters. However, some Arabic characters are wider than others. Therefore, the aspect ratio (height/width ratio) of the character is a useful feature.

### 2.5.4 Feature Normalization

The attribute data which might have different ranges (min to max) is scaled to fit in a specific range [0, 1]. We use Min–max normalization method [48] for normalization.

## 2.6 Classification

We use The Feed Forward Neural Network (back propagation) [49] which is considered one of the most powerful classifiers. The number of hidden units of ANN should be selected to be high enough to model the problem at hand but not too high to overfit. The number of hidden units is selected to have the best performance on the validation set.

### 2.6.1 Network Architecture

The neural network needs 133 inputs and 28 neurons in its output layer to identify the letters.

The network is a two-layer network. The log-sigmoid [50] activation function at the output layer was picked because its output range (0 to 1) is perfect for learning to output Boolean values. The function generates outputs between 0 and 1 as the neuron's net input goes from negative to positive infinity see Figure 5.

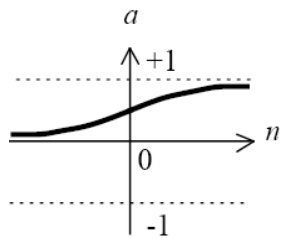

$$a = logsig(n)$$

Figure 5.  Log-sigmoid activation function.

In building the network, the data was divided randomly into two categories. Training data consisted of 80% of the data. The remaining 20% of the data was assigned to the testing data.

### 2.6.2 Network Training

To create a network that can handle noisy input vectors (characters) it is best to train the network on both ideal and noisy characters.

The network is trained to output a 1 in the correct position of the output vector and to fill the rest of the output vector with 0's.

Back propagation training method is followed. This method was selected because of its simplicity and because it has been previously used on a number of pattern recognition problems. The method uses in this work called principle of gradient descent [51].

This function is useful because the conjugate gradient algorithms have relatively modest memory requirements. Memory is important when working with large networks, and yet it is much faster than other algorithms [52].

In this paper we stop the network training is if the sum of squared error falls below 0.001.

### 2.6.3 Testing the Network

After many trials of eliminating, adding and modifying features and also adjusting network hidden layers; after 70 neurons network performance does not increase. A network of 70 neurons in hidden layer was able to predict about 88% of the input characters correctly.

## 3. RESULTS AND DISCUSSIONS

The test set size used in the experiments is 308 characters, Experimental results are presented and the best 10 recognized characters are shown in Table 1.

Table 1.  Recognition rates for best 10 Arabic characters in the proposed system.

| No | Character | Recognition Percentage |
|----|-----------|------------------------|
| 1  | خ         | 100%                   |
| 2  | ج         | 100%                   |
| 3  | ص         | 100%                   |
| 4  | ف         | 100%                   |
| 5  | ن         | 100%                   |
| 6  | و         | 100%                   |
| 7  | غ         | 99%                    |
| 8  | ه         | 97%                    |
| 9  | أ         | 96%                    |
| 10 | ل         | 96%                    |

The achieved results are very promising. Experimental results showed that the proposed method give a recognition rate of about 88% for all letters although we get success rate of 100% for some letters.

As compared to the previous works we find that Aburas [53] system achieved 70% of recognition rate, also Khedher [54] system achieved 73.4% of recognition rate, Taani [55] system achieved 75.3% of recognition rate. We achieved 1% increase in recognition over the Abandah [56] system. All of those systems work on isolated handwritten Arabic characters.

In this work Statistical and structural features were combined from the main body or the boundary with other features to get high recognition accuracy.

These features include the letter form, secondary type (Hamza,dots) and number of those secondary components. The results shown in the previous table illustrate that higher recognition accuracies are achieved using the proposed feature extraction technique. Extracting features from the main body and secondary components provides more valuable features that exploit the recognition potential of the secondary components of handwritten Arabic letters (comparison between س and ش, ص and ض). These results also confirm the importance of the secondary components of the handwritten Arabic letters.

After many trials of eliminating, adding and modifying features and also adjusting neural network hidden layers

Our database of handwritten Arabic samples is CENPRMI [57] dataset. It includes Arabic off-line isolated handwritten characters. The database contains 11620 characters (about 332 samples for each character). These characters were written according to 12 different templates by 13 writers, with each template adopted by 5–8 writers. The 11620 character samples are divided into three groups 198, 67 and 67 for training, validating and testing purposes respectively.

We found that the Recognition rate is between 100% for easy to recognize letter forms and 61% for the hardest letter forms. Table 2 shows the ten letter forms that have the lowest classification rate.

Table 2. Worst 10 recognized characters.

| No | Character | Recognition Rate |
|----|-----------|------------------|
| 1 | ق | 61% |
| 2 | ع | 66% |
| 3 | ت | 69% |
| 4 | ض | 72% |
| 5 | ث | 76% |
| 6 | س | 78% |
| 7 | ز | 80% |
| 8 | ك | 81% |
| 9 | د | 83% |
| 10 | ظ | 83% |

These ten letters are always drawn with loops or drawn with loops in some writing variations. There are substantial similarities among multiple Arabic letter form groups that have loops. Often the sole difference between such letters is a subtle difference in the loop's shape. Moreover, letters with secondaries tend to have low recognition accuracies because the variations in drawing the dots or hamzas give inaccuracies in extracting the secondary type feature. Moreover that multiple dots in some writing styles may be isolated (Naskh writing style) or continued dash (Rekaa writing style). After careful examination of the samples that were incorrectly recognized, we concluded that most of these samples are hard to recognize even by a human expert reader. However, we think that the door is open to search for extracting new features that capture subtle differences in loop shapes and secondary types.

## 4. CONCLUSION

This paper presents an approach for extracting features to achieve high recognition accuracy of handwritten Arabic characters.

We tune the used parameters during the preprocessing phase including binarization, normalization and some noise removal methods accurately to preserve all useful information that can be extracted from the character. Our algorithm extracts useful features not only from the main body, but also from the secondary components of the character. It also overcomes some of handwritten characters variations.

Selecting proper features for recognizing handwritten Arabic characters can give better recognition accuracies, therefore our feature included statistical, morphological and topological features. We pay high attention for character secondaries which enable us to distinguish between characters as this is the only way to distinguish between some characters.

Although, there are some challenges with some characters, but the overall recognition rate is perfect especially when compared to other handwritten Arabic characters systems.

After examining the recognition accuracy of each character we found that the recognition rate is between 100% for the easiest recognized characters such as (و, ن and ص) and 61% and 66% for the hardest recognized characters such as (ق and ع), respectively.

We think that this is because those characters contain loops and also they can have different characteristics from one to other writing style. We hope also that we complete system for recognizing handwritten Arabic text passing through segmentation techniques for words into characters.

# 5. REFERENCES

[1]   M. S. Khorsheed, Automatic recognition of words in Arabic manuscripts, PhD thesis, University of Cambridge, 2000.

[2]   N. Arica, Yarman-Vural. F, Optical Character Recognition for Cursive Handwriting, IEEE Trans Pattern Anal Mach Intell, 24(6), 2002, pp 801-813.

[3]   B. Al-Badr and Mahmoud, S. A, Survey and bibliography of Arabic optical text recognition. Signal Process. 41, 1, 1995, pp 49-77.

[4]   A. Nazif, A system for the recognition of the printed Arabic characters. Master's thesis, Faculty of Engineering, Cairo University, 1975.

[5]   V. Margner and H. El Abed, Databases and competitions: strategies to improve Arabic recognition systems. In Arabic and Chinese Handwriting Recognition, Lecture Notes in Computer Science, vol. 4768, Springer, 2008, pp 82-103.

[6]   M. Cheriet, Visual recognition of Arabic handwriting: challenges and new directions. In Arabic and Chinese Handwriting Recognition, Lecture Notes in Computer Science, vol. 4768, Springer, 2008, pp 1-21.

[7]   G. A. Abandah and Tareq M. Malas, "Feature Selection for Recognizing Handwritten Arabic Letters", Dirasat Engineering Sciences Journal, Vol. 37, No. 2, Oct 2010, .

[8]   S. Abdelazeem and E. EL-Sherif, "Arabic handwritten digit recognition," international Journal on Document Analysis and Recognition (IJDAR),11, 3, , 2008, pp 127–141.

[9]   G. A. Abandah and N. Anssari, "Novel moment features extraction for recognizing handwritten Arabic letters," Journal of Computer Science, 5, 3, 2009, pp 226-232.

[10]  A. Amin, H. Al-Sadoun and S. Fischer, Hand- Printed Arabic Character Recognition System Using an Artificial Network. Pattern Recognition, 29, 1996, pp 663-675.

[11]  A. Amin, Recognition of Hand-Printed Characters Based on Structural Description and Inductive Logic Programming. Pattern Recognition Letters, 24, 2003,  pp 3187-3196.

[12]  G. Olivier, H. Miled, K. Romeo & Y. Lecourtier, Segmentation and Coding of Arabic Handwritten Words. Proc. 13th Int'l Conf.Pattern Recognition, 3, 1996,  pp 264-268.

[13]  I.S.I. Abuhaiba and P. Ahmed, Restoration of Temporal Information in Off-Line Arabic Handwriting, Pattern Recognition, 26, 1993, pp 1009-1017.

[14]  I.S.I. Abuhaiba, S.A. Mahmoud and R.J. Green, Recognition of Handwritten Cursive Arabic Characters. IEEE Trans. Pattern Analysis and Machine Intelligence, 16, 1994, pp 664-672.

[15]  M. Dehghan, K.Faez, M. Ahmadi and M. Shridhar, Handwritten Farsi (Arabic)Word Recognition: A Holistic Approach Using Discrete HMM. Pattern Recognition, 34, 2001, pp 1057-1065.

[16]  H. Al-Yousefi and S.S. Udpa, Recognition of Arabic Characters. IEEE Trans. Pattern Analysis and Machine Intelligence, 14, 1992, pp 853-857.

[17]  S. Abdelazeem and E. EL-Sherif, Arabic handwritten digit recognition. Int. J. Doc. Anal. Recog. 11, 3, 2008, pp 127-141.

[18]  G. Abandah and N. Anssari, Novel moment features extraction for recognizing handwritten Arabic letters. J. Comput. Sci. 5, 3, 2009, pp 226-232.

[19]  M. Kherallah, L. Haddad, A. M. Alimi, and A. Mitiche, "On-Line handwritten digit recognition based on trajectory and velocity modeling", Pattern Recogn. Lett. 29, 5, 2008, pp 580-594.

[20]  M. Kherallah, F. Bouri, and A. M. Alimi, "On-line Arabic handwriting recognition system based on visual encoding and genetic algorithm" Engin. Appl. Artif. Intell. 22, 1, 2009, pp 153-170.

[21]  N. N. Mezghani, and A. Mitiche, A gibbsian kohonen, "network for online arabic character recognition". In Advances in Visual Computing, Lecture Notes in Computer Science, vol. 5359, Springer, 2008, pp 493-500.

[22]    R. Saabni, and J. El-sana, "Hierarchical on-line Arabic handwriting recognition". In Proceedings of the 10th International Conference on Document Analysis and Recognition (ICDAR), 2009, pp 867-871.

[23]    J. Sternby, J. Morwing, J. Andersson, and C. FRriberg, "On-Line Arabic handwriting recognition with templates", Pattern Recogn., New Frontiers Handwrit. Recogn. 42, 12, 2009, pp 3278-3286.

[24]    M. Benjelil, S.Kanoun, R.Mullot and A. M. Alimi, "Arabic and latin script identification in printed and handwritten types based on steerable pyramid features", In Proceedings of the Proceedings of the 10th International Conference on Document Analysis and Recognition (ICDAR), 2009, pp 591-595.

[25]    I.B.Cheikh, A.Bela, and A.Kacem, "A novel approach for the recognition of a wide Arabic handwritten word lexicon", In Proceedings of the 19th International Conference on Pattern Recognition (ICPR), 2008,  pp *1-4*.

[26]    S. Kanoun, F.Slimane, H.Guesmi, R.Ingold, A. M.Almi, and J.Hennebert, Affixal approach versus analytical approach for off–line arabic decomposable vocabulary recognition. In Proceedings of the 10th International Conference on Document Analysis and Recognition (ICDAR), 2009, pp 661-665.

[27]    T. K. Khan, S. M. Azam and S. Mohsin, "An improvement over template matching using k-means algorithm for printed cursive script recognition", In Proceedings of the 4th IASTED International Conference on Signal Processing, Pattern Recognition, and Applications,  2007, pp 209-214.

[28]    S. Benmoussa , Q. Frissard, A. Zahour, A. Benabdelhafid and A. M. Alimi, "New features using fractal multi-dimensions for generalized Arabic font recognition", Pattern Recogn. Lett. 31, 5, 2010, pp 361-371.

[29]    R.Prasad, S. Saleem, M. Kamali, R. Meermeier and P. Natarajan, "Improvements in hidden markov model based Arabic ocr",  In Proceedings of the 19th International Conference on Pattern Recognition (ICPR). 2008.

[30]    K.  Saeeda and M. Albakoor, "Region growing based segmentation algorithm for typewritten and handwritten text recognition" Appl. Soft Comput. 9, 2, 2009. pp 608–617.

[31]    F. Slimane, R. Ingold, S. Kanoun, A. M. Alimi, and J. Hennebert, A new Arabic printed text image database and evaluation protocols. In Proceedings of the 10th International Conference on Document Analysis and Recognition (ICDAR), 2009, pp 946–950.

[32]    G. Abandah and N. Anssari, Novel moment features extraction for recognizing handwritten Arabic letters. J. Comput. Sci. 5, 3, 2009, pp 226-232.

[33]    K. Sherif and M. Mostafa,  A Parallel Design and Implementation For Backpropagation Neural Network Using MIMD Architecture. IEEE, 1996, pp 1361- 1366.

[34]    K. Sherif & M. Mostafa,  A Parallel Design and Implementation For Backpropagation Neural Network Using MIMD Architecture. IEEE, 1996, pp 1472- 1475.

[35]    M. A. Ali, Arabic handwritten characters classification using learning vector quantization algorithm. In Image and Signal Processing, Lecture Notes in Computer Science, vol. 5099, Springer, 2008, pp 463-470.

[36]    T. Bluche, H. Ney, C. Kermorvant, "Feature extraction with convolutional neural networks for handwritten word recognition," 12th International Conference on Document Analysis and Recognition, 2013, pp 285-289.

[37]    M. Kozielski, P. Doetsch and H. Ney, "Improvements in RWTH's system for off-line handwriting recognition," 12th International Conference on Document Analysis and Recognition, 2013, pp 935-939.

[38]    L. Rothacker, S. Vajda, and G. A. Fink, "Bag-of-features representations for offline handwriting recognition applied to Arabic script," in Proceedings of the 3rd International Conference on Frontiers in Handwriting Recognition (ICFHR '12), Bari, Italy, 2012, pp 149–154.

[39]    M. Kozielski, P. Doetsch and H. Ney, "Improvements in RWTH's system for off-line handwriting recognition," 12th International Conference on Document Analysis and Recognition, 2013, pp 935-939.

[40]    S. A. Mahmoud and S. O. Olatunji, "Automatic recognition of off-line handwritten Arabic (Indian) numerals using support vector and extreme learning machines," International Journal of Imaging 2, A09, 2009.

[41]    S. A. Mahmoud and S. Owaidah, "Recognition of off-line handwritten Arabic (Indian) numerals using multi-scale features and support vector machines," Arabian Journal for Science & Engineering (Springer), Vol. 34 Issue 2B, 2009, pp 429-444.

[42]    M. Gargouri, S. Kanoun and J.M. Ogier, "Text-independent Writer Identification on Online Arabic Handwriting," 12th International Conference on Document Analysis and Recognition, 2013, pp 428-432.

[43]    Otsu, N., "A Threshold Selection Method from Gray-Level Histograms," IEEE Transactions on Systems, Man, and Cybernetics, Vol. 9, No. 1, 1979, pp 62-66.

[44]    G. Lazzara and T. Géraud, "Efficient multiscale Sauvola's binarization," Springer-Verlag Berlin Heidelberg, 2013.

[45]    Cheriet. M, Character Recognition Systems, John Wiley, 2007, 36-38.

[46]    S.L. Jae, Two dimensional signal and image processing, PRENTICE HALL PTR, Upper Saddle River, New Jersey. 07458, 1990.

[47]    A. Rosenfeld, and A. Kak, "Digital Picture Processing," 1st Edn, Academic Press, New York, ISBN: 10: 0125973608, 1976.

[48]    J. Hann, M.Kamber,. Data Mining: Concepts and Techniques, 3rd Edition. Morgan Kaufman Publishers, 2011.

[49]    S. Daniel, K. Vladimír, P. Jiri, Introduction to multi-layer feed-forward neural networks, ELSEVIER, Chemometrics and Intelligent Laboratory Systems 39, 1997, pp 43-62.

[50]    P. CHANDRA, Sigmoidal Function Classes for Feedforward Artificial Neural Networks, Kluwer Academic Publishers, 2003, pp 185-195.

[51]    M. F. Moller, A scaled conjugate gradient algorithm for fast supervised learning, Neural Networks, Vol. 6, 1993, pp 525-533.

[52]    M. H. Beale, T. Hagan and B. Demuth, Neural Network Toolbox 7, User's Guide, by The MathWorks, Inc., 2010.

[53]    A. A. Aburas and S. M. Rehiel, Off-line Omni-style Handwriting Arabic Character Recognition System Based on Wavelet Compression, Arab Research Institute in Sciences & Engineering ARISER, 2007, pp 123-35.

[54]    M. Z. Khedher, G. A. Abandah, and A. M. Al-Khawaldeh, Optimizing Feature Selection for Recognizing Handwritten Arabic Characters, World Academy of Science, Engineering and Technology 4, 2005, pp 81-84.

[55]    A. T. Al-Taani and S. Al-Haj, Recognition of On-line Arabic Handwritten Characters Using Structural Features, Journal of Pattern Recognition Research, 2010, pp 23-37.

[56]    G. A. Abandah, K. S. Younis and M. Z. Khedher, Handwritten Arabic Character Recognition Using Multiple Classifiers Based On Letter Form, Conf. on Signal Processing, Pattern Recognition, & Applications Austria, 2008, pp 128-133.

[57]    H. Alamri, J.Sadri, C.Y.Suen, N.Nobile, "A novel comprehensive database for Arabic off line handwriting recognition," The11thInternational Conference on Frontiers in Handwriting Recognition (ICFHR), 2008, pp 664–669.

## Authors

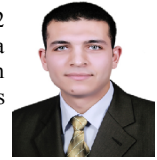

**Ahmed T. Sahlol** Obtained Bachelor degree from Mansoura University, Egypt in 2004, 2 years Diploma from Mansoura University in 2006, Master degree from Mansoura University, Egypt in 2010. He is currently visiting researcher at Concordia University in Canada and he got a governmental scholarship to study his PhD. His research interests include pattern recognition, optical character recognition and image processing.

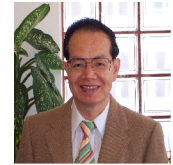

**Ching Y. Suen** is the Director of CENPARMI and the Concordia Chair on AI and Pattern Recognition. He has served as the Chairman of the Department of Computer Science and as the Associate Dean (Research) of the Faculty of Engineering and Computer Science of Concordia University. His research projects have been funded by the ENCS Faculty and the Distinguished Chair Programs at Concordia University, FCAR (Quebec), NSERC (Canada), the National Networks of Centres of Excellence (Canada) and the industrial sectors in various countries, including Canada, France, Japan, Italy, and the United States.